\documentclass[graybox]{styles/svmult}

\usepackage{mathptmx}       %
\usepackage{helvet}         %
\usepackage{courier}        %
\usepackage{type1cm}        %
\usepackage{makeidx}         %
\usepackage{graphicx}        %
\usepackage{multicol}        %
\usepackage[bottom]{footmisc}%

\usepackage{amsmath,amssymb,amsfonts}
\usepackage{graphicx}
\usepackage{textcomp}
\usepackage{xcolor}
\usepackage{comment}
\usepackage[caption=false,font=footnotesize,labelfont=sf,textfont=sf]{subfig}
\usepackage[colorlinks=true, urlcolor=blue, citecolor=blue, linkcolor=darkgray]{hyperref}

\def\BibTeX{{\rm B\kern-.05em{\sc i\kern-.025em b}\kern-.08em
    T\kern-.1667em\lower.7ex\hbox{E}\kern-.125emX}}

\makeindex             %

\begin{document}

\title*{Place-Revisit Consistency Evaluation of Reference Poses}
\author{Jianzhu Huai$^{1}$, Yujia Zhang*$^{2, }$$^{3}$, and Grzegorz Jozkow$^{4}$}

\maketitle
\begin{center}
\small
\vspace{-3.0cm}

$^{1}$ State key Lab of Info Engineering in Surveying, Mapping and Remote Sensing (LIESMARS), Wuhan University, Wuhan, Hubei, China (jianzhu.huai@whu.edu.cn)

$^{2}$ School of Instrument Science and Engineering, State Key Laboratory of
Comprehensive PNT Network and Equipment Technology, Southeast University, Nanjing, Jiangsu, China ({yujia\_zhang@seu.edu.cn})

$^{3}$ UK-China Intelligent Mobility Research Centre, Southeast University, Nanjing, Jiangsu, China

$^{4}$ Institute of Geodesy and Geoinformatics, 
Wroclaw University of Environmental and Life Sciences, Wroclaw, Poland (grzegorz.jozkow@upwr.edu.pl)

\vspace{0.3em}

$^{*}$ Corresponding author:
\href{mailto:yujia_zhang@seu.edu.cn}{yujia\_zhang@seu.edu.cn}

\end{center}

\vspace{0.8em}

\abstract{ Reference trajectories are widely used to evaluate odometry and simultaneous localization and mapping systems. However, local inconsistencies in benchmark reference poses can bias such evaluations. This paper presents a place-revisit consistency evaluation method for reference poses in mobile sensing datasets. When a platform revisits a previously observed place after a sufficient temporal gap, the relative pose implied by the reference trajectory should agree with relative poses estimated directly from onboard sensors. We instantiate this idea with lidar scan registration and stereo-image pose estimation, and apply it to the KITTI Odometry and SNAIL Radar. The KITTI analysis identifies several revisit regions where the public reference poses disagree with sensor-derived relative poses, with median translation discrepancies up to 6.60 m. Representative point-cloud overlays further localize these discrepancies and support the use of place revisits as a benchmark-validation tool.}

\section{Introduction}
\label{sec:1}
Reference trajectories are central to odometry and SLAM benchmark evaluation, 
where trajectory-level error metrics are commonly used to compare systems~\cite{sturmBenchmarkEvaluationRGBD2012}. 
They are widely treated as ground truth for evaluating visual odometry, lidar odometry, and SLAM systems. 
However, reference trajectories are not strictly error-free. 
They are produced through sensor processing pipelines and calibration procedures that inevitably introduce systematic errors.
In many benchmarks, the accuracy characteristics of the reference poses are not fully documented, which limits the ability to assess their reliability.
A high-quality benchmark should ideally provide reference trajectories that are demonstrably more accurate than the methods being evaluated;
otherwise, evaluation results may be biased and the benchmark itself becomes less trustworthy.

This paper studies reference-pose consistency with a key observation.
When a vehicle revisits a previously observed place after a sufficient temporal gap,
the relative pose derived from the reference trajectory should agree with the relative pose estimated directly from sensor measurements.
If independent modalities such as lidar scan matching and stereo geometry produce consistent relative poses that disagree with the reference trajectory, 
this indicates a potential local inconsistency in the reference poses rather than a failure of a specific sensing method.

The proposed analysis is related to prior work on odometry benchmarking, scan registration, and visual relative-pose estimation.
Unlike conventional odometry evaluation, which assumes that the benchmark trajectory is correct,
we use independent sensor-derived relative poses to evaluate the reference trajectory itself.
The method is particularly suited to repeated-route or looped mobile-sensing datasets where place revisits occur naturally.
The principal contributions are summarized in three aspects. 
First, we formulate a place-revisit consistency check for reference poses in mobile sensing benchmarks. 
Second, we instantiate the check using lidar Generalized ICP and stereo PnP as independent relative-pose estimates. 
Third, we apply the method to KITTI Odometry sequences 00--10 and identify revisit regions with multi-meter reference-pose discrepancies.
We also evaluate lidar scans and stereo images from the SNAIL Radar dataset, with motion compensation applied to the lidar scans.

\section{Place-Revisit Consistency Evaluation}
\label{sec:2}
\subsection{Candidate Revisit Selection}
To check a sequence, the input data consist of the reference poses and the lidar scans. If stereo images are to be used, we also need the camera-lidar calibration.
For instance, with a KITTI sequence, the input data consist of the benchmark pose file, velodyne scans, stereo gray images, and the stereo camera-lidar calibration.
We first convert KITTI left-camera reference
poses $\{\mathbf T_{w \leftarrow cam}\}$ to lidar poses $\{\mathbf T_{w \leftarrow lidar}\}$ using the calibration transform
\(\mathbf T_{cam\leftarrow lidar}\). 
We use the convention $\mathbf T_{i\leftarrow j}$, 
which maps points from frame $j$ into frame $i$.
And $\mathbf{T}_{i\leftarrow j}$ consists of a rotation matrix $\mathbf{R}_{i\leftarrow j} \in SO(3)$ 
and a translation vector $\mathbf{t}_{i\leftarrow j}$.

Candidate revisit pairs are selected using temporal and spatial constraints on the reference positions $\mathbf{p}^{ref}$,
\begin{equation}
    j-i > \tau_t,\quad
    \left\|\mathbf{p}^{ref}_{w\leftarrow i}-\mathbf{p}^{ref}_{w\leftarrow j}\right\|_2 \leq \tau_s,
\end{equation}
where $i$ is the train frame, $j$ is the query frame, $\tau_t$ is the minimum temporal gap, and $\tau_s$ is the maximum spatial distance.
Dataset-specific selection thresholds are given in Section~\ref{sec:experiments}.

\subsection{Reference and Sensor-Derived Relative Poses}
For each candidate pair, the reference relative pose is
\begin{equation}
\mathbf T^{\mathrm{ref}}_{i\leftarrow j}=\left(\mathbf T^{\mathrm{ref}}_{w\leftarrow i}\right)^{-1}\mathbf T^{\mathrm{ref}}_{w\leftarrow j}.
\label{eq:ref_relative}
\end{equation}

The lidar estimate $\mathbf T^{\mathrm{lidar}}_{i\leftarrow j}$ is obtained by registering the query scan to the train scan with Generalized ICP \cite{segalGeneralizedICP2009}. The reference relative pose in \eqref{eq:ref_relative} is used as the initial guess. Registration convergence, fitness, point counts, and aligned point clouds are recorded for validity filtering and inspection.

The stereo estimate $\mathbf T^{\mathrm{stereo}}_{i\leftarrow j}$ is obtained from feature matching, stereo triangulation and PnP. SIFT features are matched between train left/right images for stereo geometry and between train-left and query-left images for temporal association. Triangulated train-frame landmarks and query-left image points are passed to PnP with RANSAC, refined on inliers, and converted to the lidar-frame convention using the camera-lidar calibration.

\subsection{Consistency Metrics}
For a sensor-derived estimate $\mathbf T^{\mathrm{est}}_{i\leftarrow j}$, the translation discrepancy relative to the reference pose is
$e_t=\left\|\mathbf{t}^{\mathrm{est}}_{i\leftarrow j}-\mathbf{t}^{\mathrm{ref}}_{i\leftarrow j}\right\|_2$,
and the rotation discrepancy is
\begin{equation}
e_R=\mathrm{angle}\left(\left(\mathbf R^{\mathrm{ref}}_{i\leftarrow j}\right)^\top \mathbf R^{\mathrm{est}}_{i\leftarrow j}\right)
\end{equation}
where angle($\cdot$) denotes the rotation angle obtained from the axis–angle representation of a rotation matrix.

For each train frame $i$, there may exist multiple valid revisit candidates $j$. Each candidate pair produces a translation error 
$e_t(i,j)$ and rotation error $e_R(i,j)$.
To obtain a robust per-frame summary, we compute the median error over all valid query frames associated with the same train frame,
$\widetilde{e}_t(i) = \mathrm{median}_j e_t(i,j)$,
$\widetilde{e}_R(i) = \mathrm{median}_j e_R(i,j)$.
For a train frame $i$, the ``median representative'' is the specific candidate pair $(i,j)$ whose error is closest to the median,
$j^* = \arg \mathrm{min} |e_t(i,j) - \widetilde{e}_t(i)|$.

We also identify the translation axis with the largest accumulated absolute lidar/reference discrepancy,
\begin{equation}
a^\star=\arg\max_{a\in\{x,y,z\}}\sum_k\left|\mathbf{t}^{\mathrm{lidar}}_{k,a}-\mathbf{t}^{\mathrm{ref}}_{k,a}\right|.
\end{equation}

\section{Experiments}
\label{sec:experiments}
\subsection{KITTI Odometry}

We evaluate the proposed place-revisit consistency analysis on KITTI Odometry sequences 00--10, with $\tau_t=200$ frames and $\tau_s=10$ m.
KITTI remains a widely used benchmarks for odometry and SLAM evaluation~\cite{geigerAreWeReady2012}, 
supporting research in stereo visual odometry \cite{huaiStereoinertialOdometryUsing2015}, 
lidar-based mapping \cite{koideGlobally2021, cwianMADBA3DLiDAR2025}, and visual 3D reconstruction \cite{cheng2026horizonstream}. 

Table~\ref{tab:summary} summarizes the candidate-pair counts, the number of train frames with a valid median representative, the maximum median lidar/reference discrepancies, and the dominant translation-error axis.
Sequences 01, 03, 04, and 10 produced no revisit pairs under the selected temporal and spatial thresholds.

\begin{table}[!t]
\caption{KITTI revisit-pair summary.
L/S denote the number of train frames with at least one valid lidar/stereo revisit candidate.}
\label{tab:summary}
\centering
\footnotesize
\begin{tabular}{p{1cm}p{1.2cm}p{1.2cm}p{1.2cm}p{2cm}p{2cm}p{1cm}}
\hline\noalign{\smallskip}
Seq & Pairs & L & S & Max $\widetilde{e}_t(i)$ & Max $\widetilde{e}_R(i)$ & Axis\\
 & & & & (m) & (deg.) & \\
\hline
00 & 29582 & 1045 & 936 & 1.79 & 1.70 & z\\
02 & 7113 & 491 & 338 & 4.83 & 3.51 & z\\
05 & 16371 & 608 & 516 & 0.62 & 0.83 & z\\
06 & 5461 & 308 & 294 & 0.17 & 0.90 & y\\
07 & 2893 & 49 & 33 & 0.26 & 0.31 & y\\
08 & 9247 & 398 & 7 & 6.60 & 2.78 & z\\
09 & 565 & 38 & 38 & 4.16 & 0.29 & z\\
\hline
\end{tabular}
\end{table}

\begin{figure}[!htbp]
\centering
\subfloat[]{\includegraphics[width=0.8\linewidth]{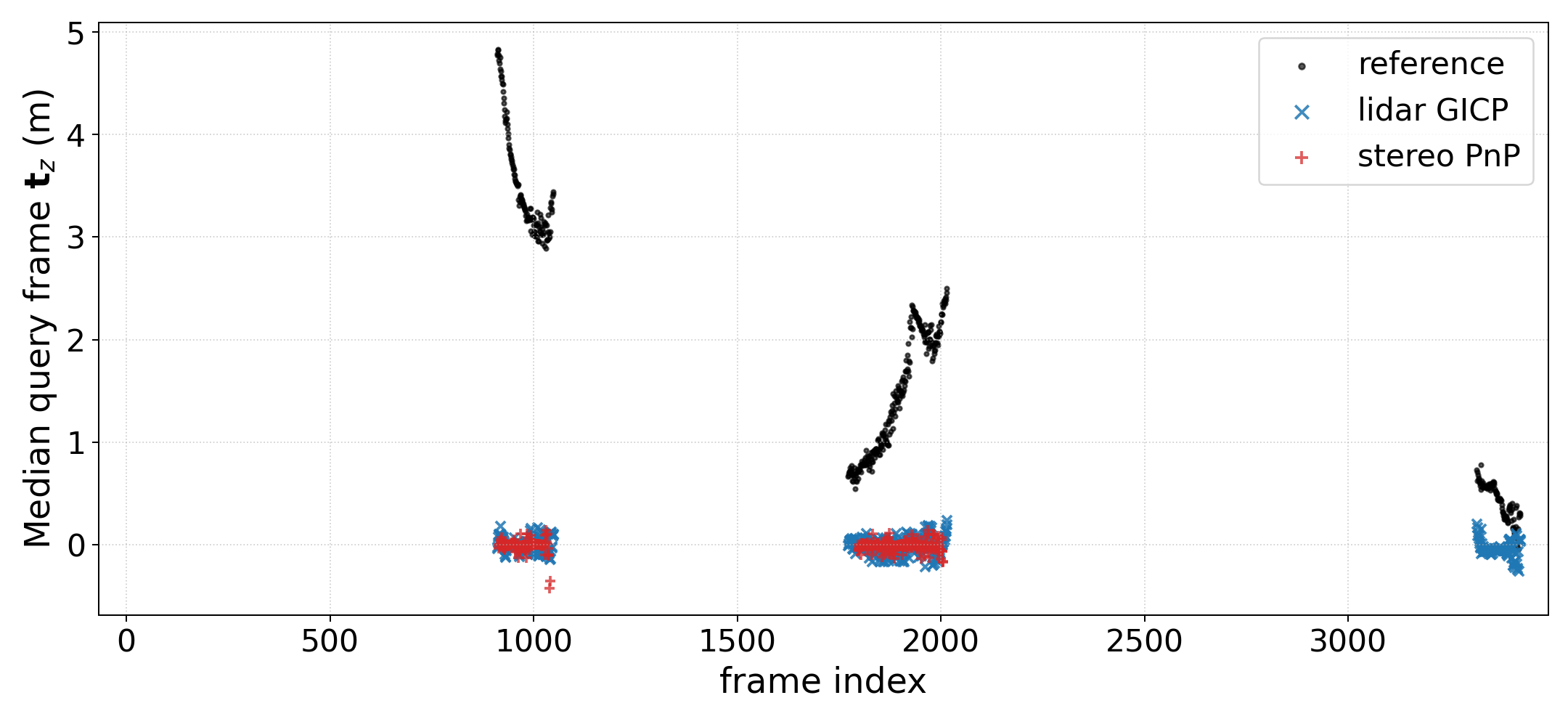}}\\[-0.4ex]
\subfloat[]{\includegraphics[width=0.8\linewidth]{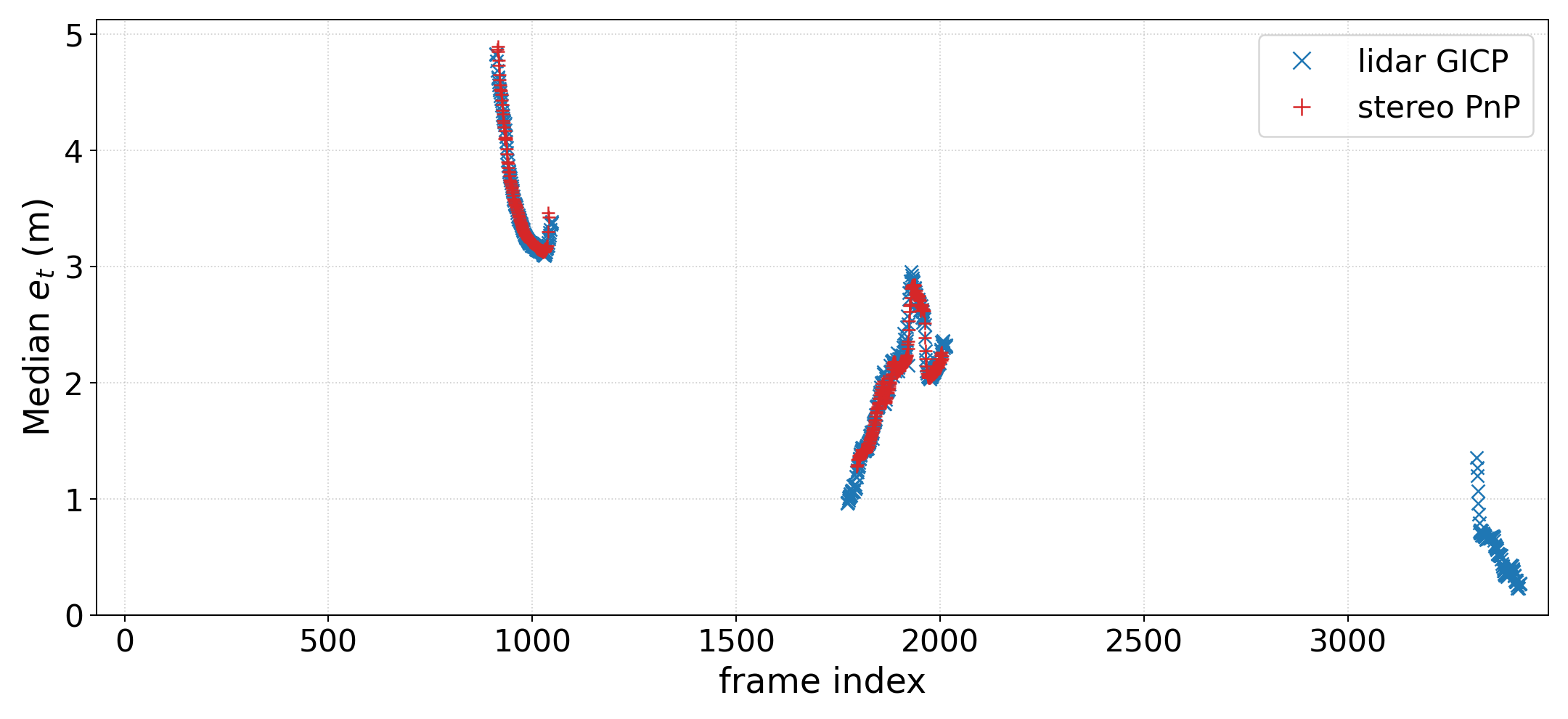}}\\[-0.4ex]
\subfloat[]{\includegraphics[width=0.8\linewidth]{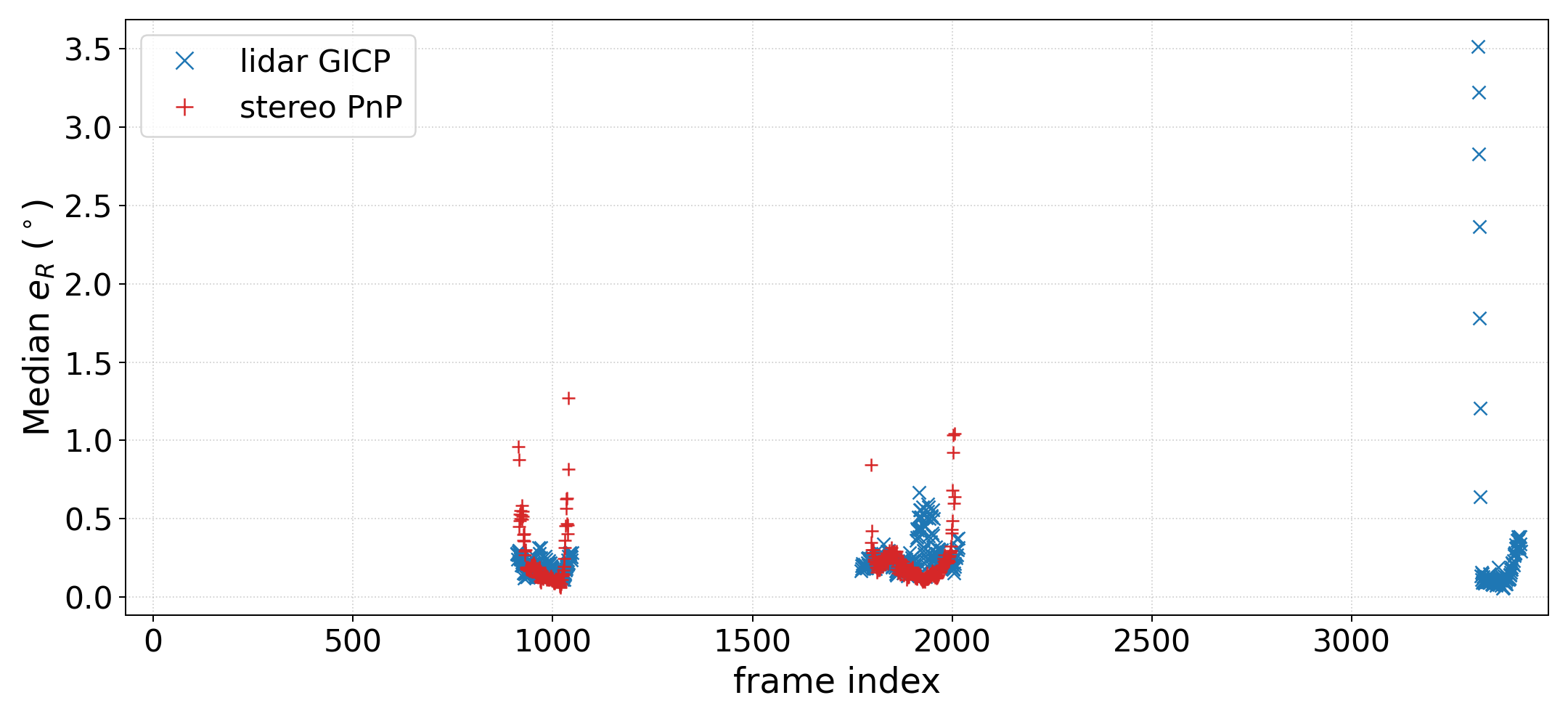}}\\[-0.4ex]
\subfloat[]{
\includegraphics[width=0.6\linewidth]{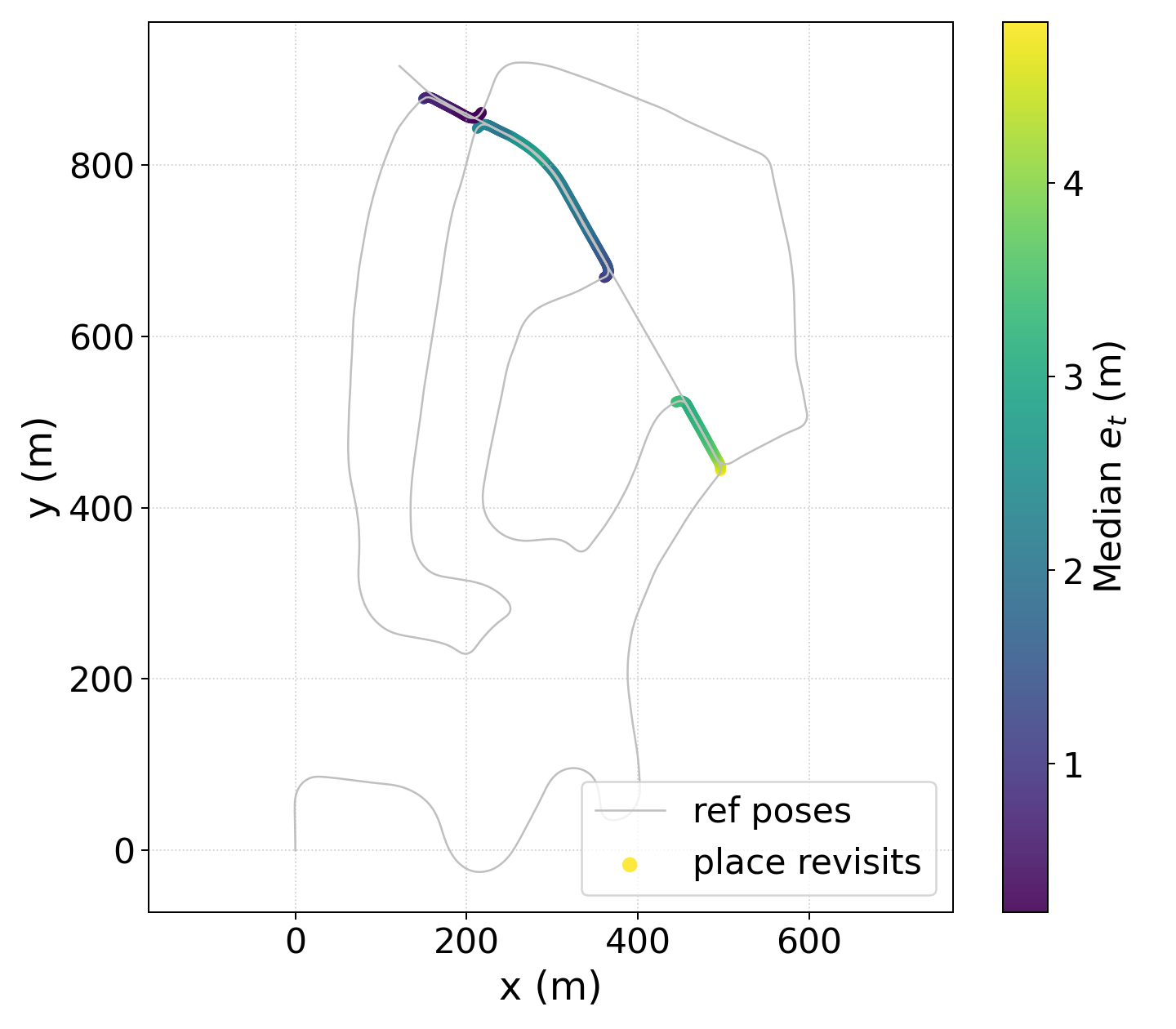}}
\caption{Sequence 02 diagnostics. (a) 
Relative translation on the dominant error axis for the median representative revisit pair selected at each train frame. (b) Median translation discrepancy. (c) Median rotation discrepancy. (d) Median lidar/reference
translation discrepancy on the reference trajectory.}
\label{fig:seq02}
\end{figure}

\begin{figure}[!htbp]
\centering
\subfloat[]{\includegraphics[width=0.8\linewidth]{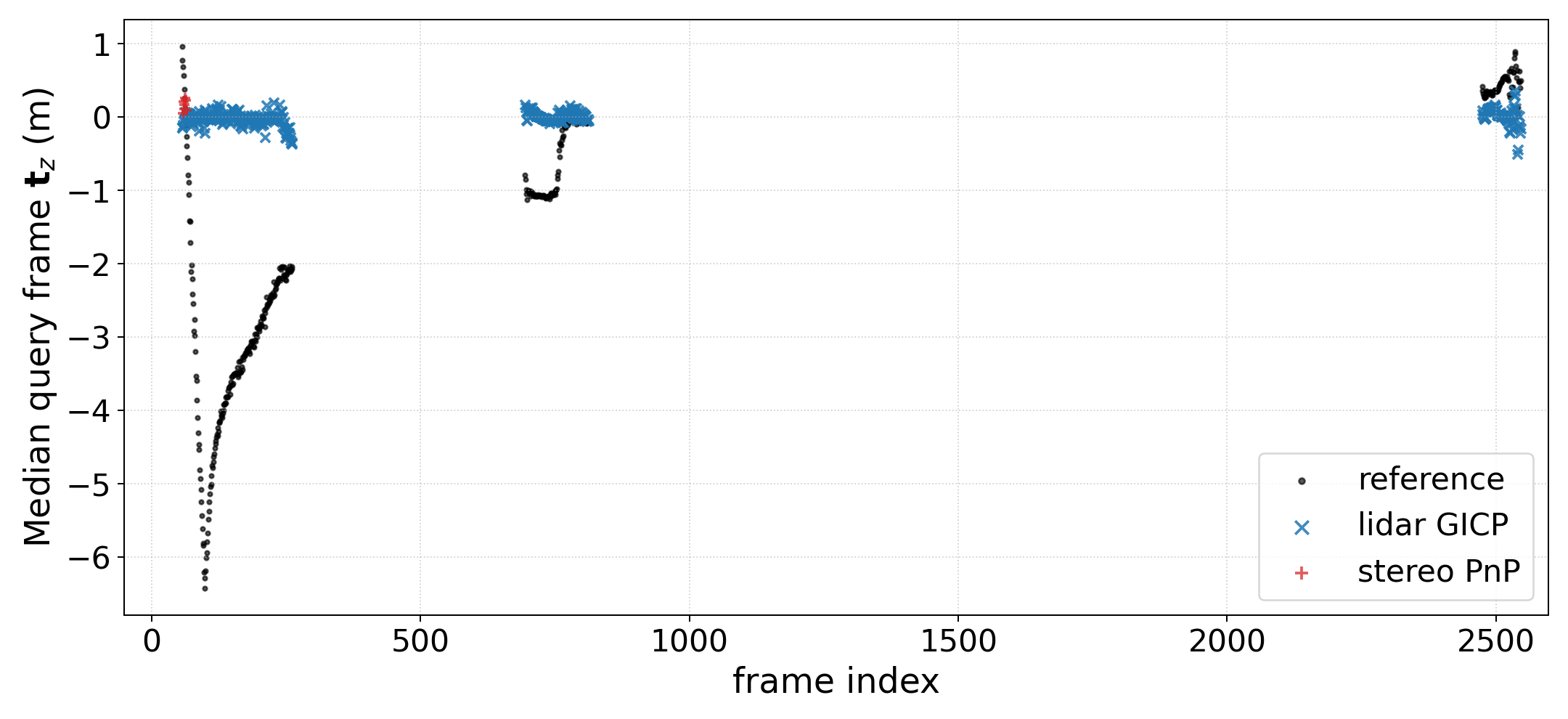}}\\[-0.4ex]
\subfloat[]{\includegraphics[width=0.8\linewidth]{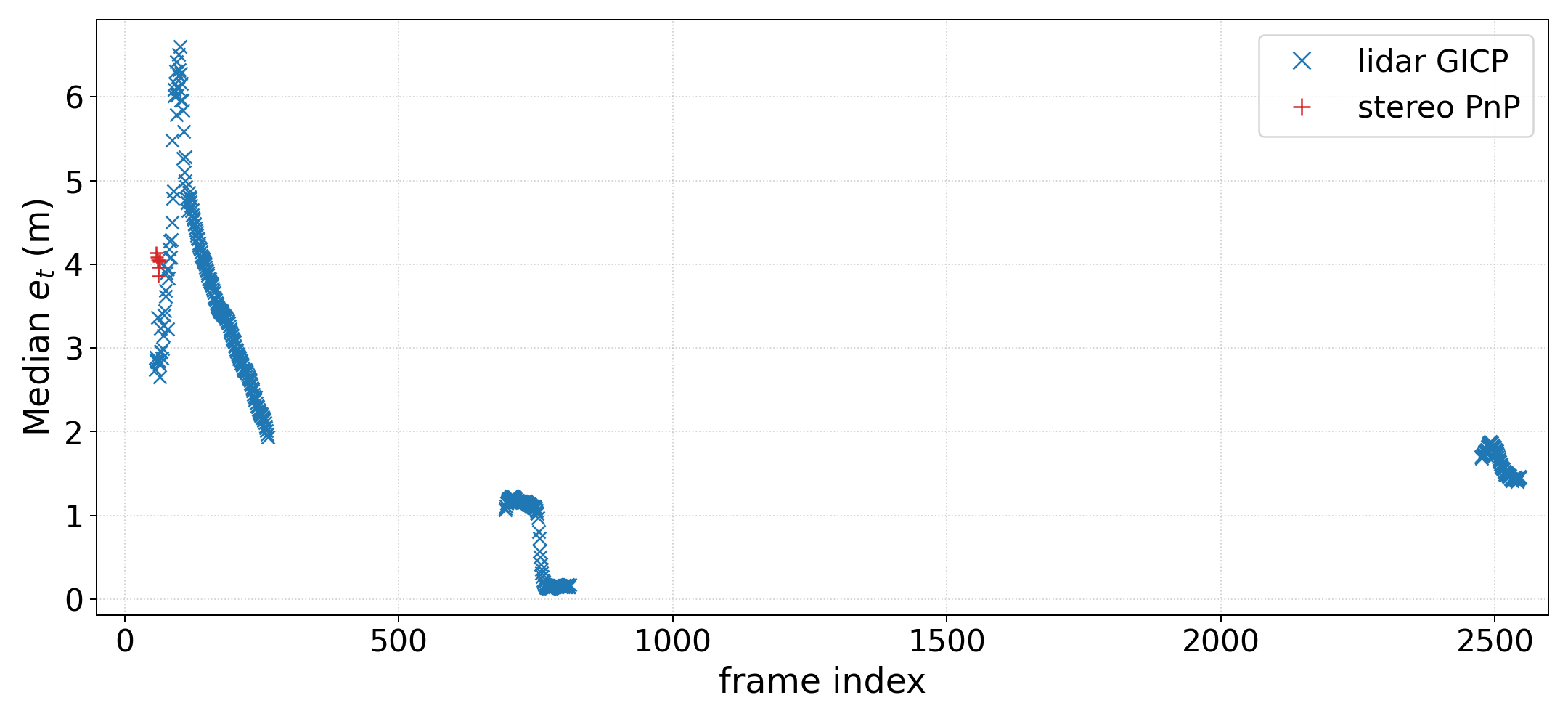}}\\[-0.4ex]
\subfloat[]{\includegraphics[width=0.8\linewidth]{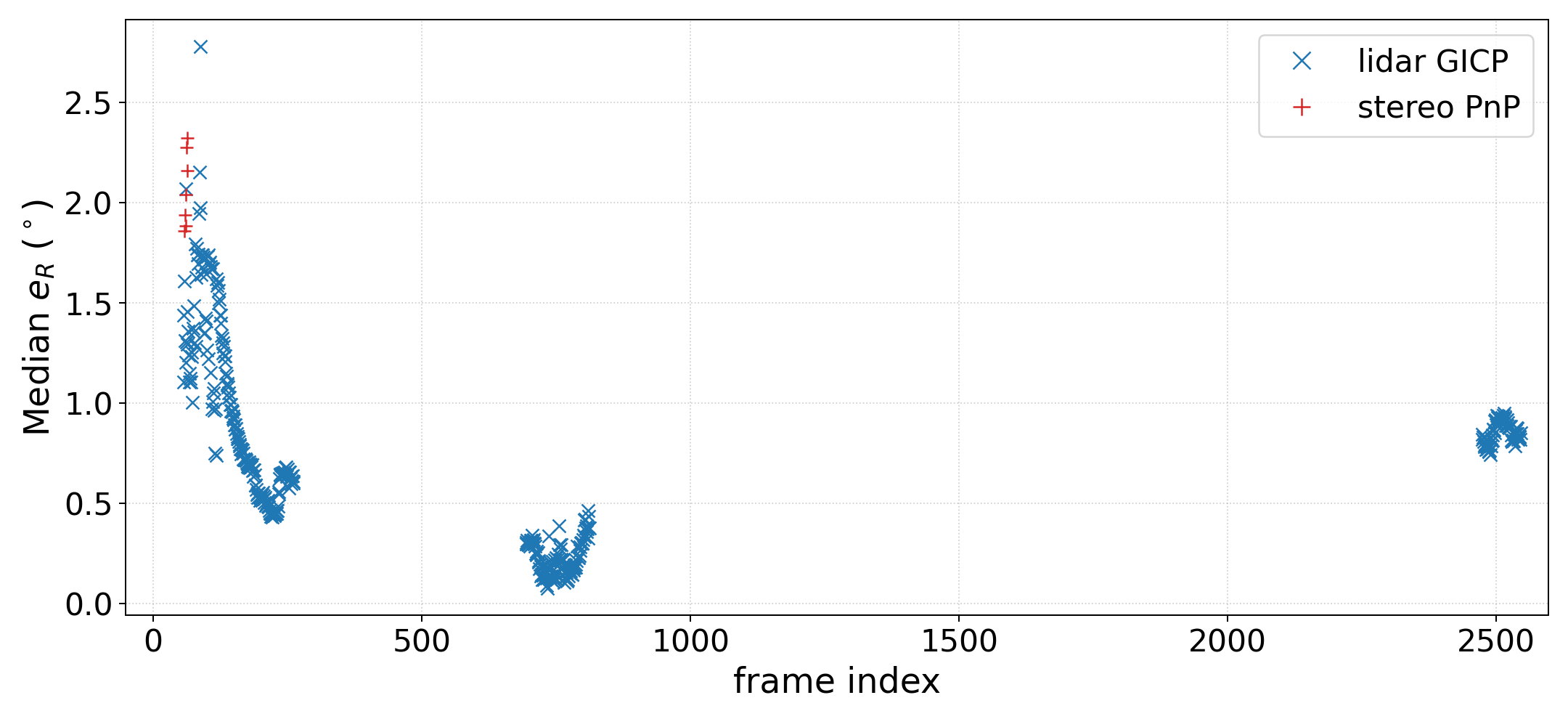}}\\[-0.4ex]
\subfloat[]{
\includegraphics[width=0.6\linewidth]{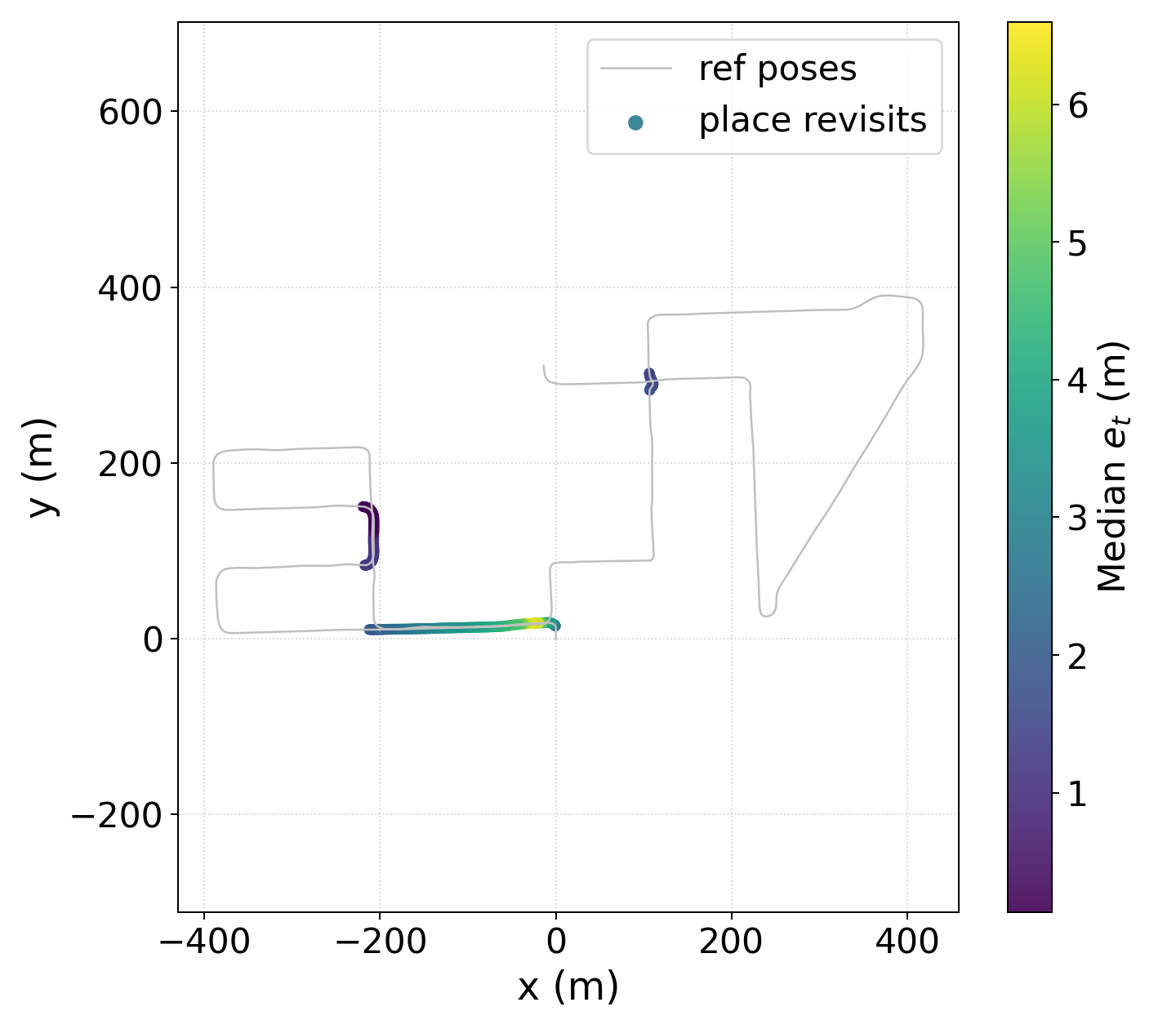}}
\caption{Sequence 08 diagnostics. (a) 
Relative translation on the dominant error axis for the median representative revisit pair selected at each train frame. (b) Median translation discrepancy. 
(c) Median rotation discrepancy.
(d) Median lidar/reference
translation discrepancy on the reference trajectory.}
\label{fig:seq08}
\end{figure}

The largest discrepancies occur in sequences 08, 02, and 09. Sequence 08 has
the largest median lidar/reference translation discrepancy, reaching 6.60 m,
while sequences 02 and 09 reach 4.83 m and 4.16 m, respectively. In contrast,
sequences 05, 06, and 07 are comparatively benign under the same analysis.
Sequence 02 (Fig.~\ref{fig:seq02})
has a substantial number of valid stereo estimates after filtering, and the
stereo relative poses mostly agree with the lidar estimates. This cross-sensor
agreement suggests that the discrepancy is unlikely to be caused by lidar
registration alone. 
In sequence 08 (Fig.~\ref{fig:seq08}), stereo estimates are sparse after filtering,
so lidar provides the main evidence; nevertheless, the lidar estimates remain
locally consistent across neighboring train frames.

\begin{figure}[!htbp]
\centering
\subfloat[]{\includegraphics[width=0.95\linewidth]{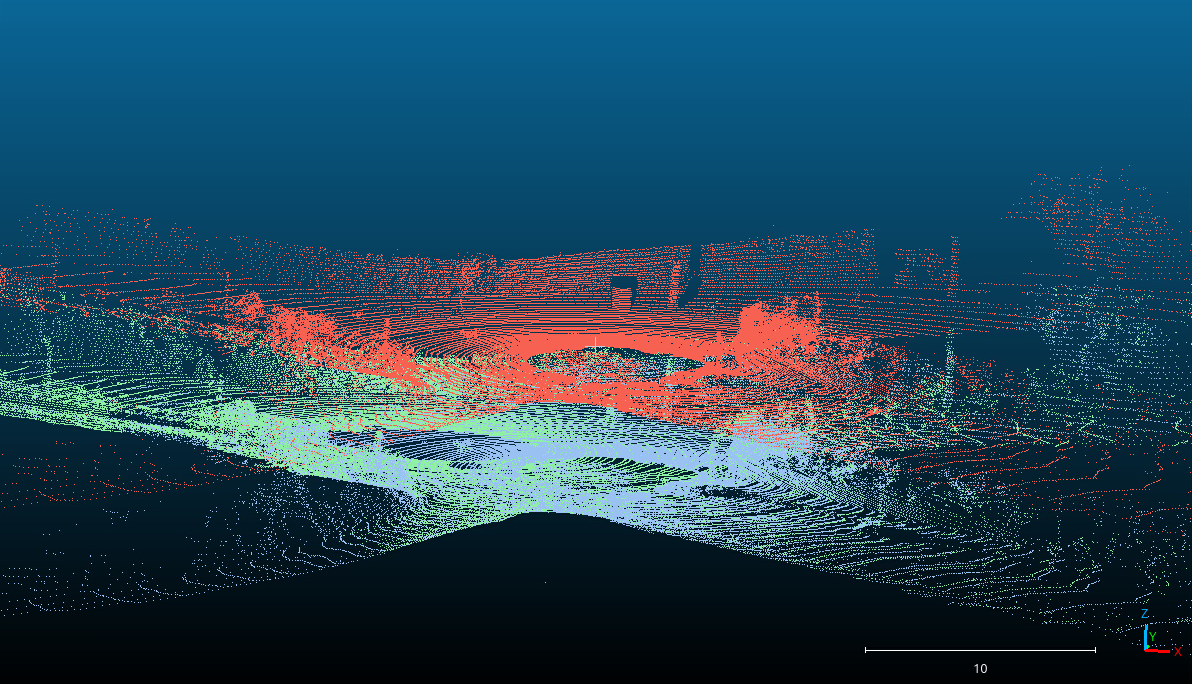}}\\[-0.3ex]
\subfloat[]{\includegraphics[width=0.95\linewidth]{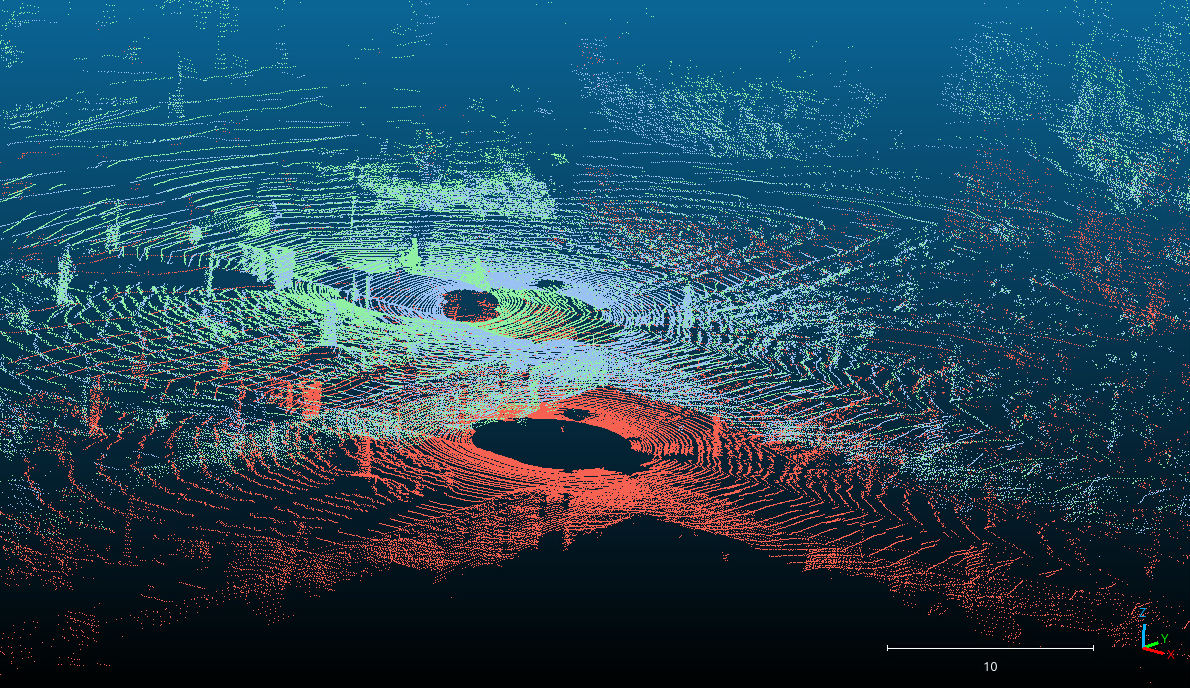}}\\[-0.3ex]
\subfloat[]{\includegraphics[width=0.95\linewidth]{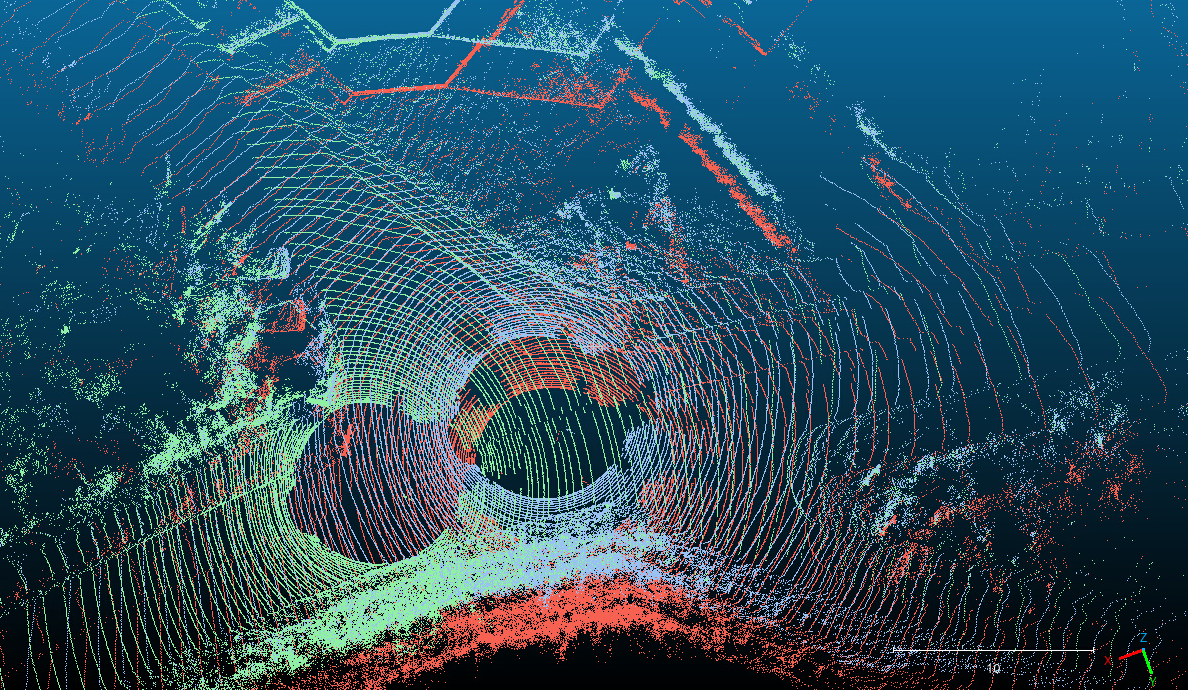}}
\caption{Point-cloud overlays for representative place revisits. Green: train frame. Blue: query frame transformed by GICP. Red: query frame transformed by the reference relative pose. (a) Sequence 02, $911 \leftarrow 4186$. (b) Sequence 08, $95 \leftarrow 1792$. (c) Sequence 09, $31 \leftarrow 1585$.}
\label{fig:overlay}
\end{figure}

We further inspect representative revisit pairs by overlaying the train point cloud with the query point cloud transformed by either the GICP-estimated relative pose or the reference relative pose. Fig.~\ref{fig:overlay} shows that the GICP-transformed query cloud aligns with the train cloud, while the reference-transformed query cloud has visible offsets. The inspected pairs are sequence 02 frames $911 \leftarrow 4186$, sequence 08 frames $95 \leftarrow 1792$, and sequence 09 frames $31 \leftarrow 1585$, with lidar/reference translation discrepancies of 4.81 m, 6.43 m, and 4.08 m, respectively.

\subsection{SNAIL Radar}
\label{sec:snail_results}
We additionally evaluate lidar and stereo data and full reference trajectory from 11 clear, daytime sequences of the SNAIL Radar dataset~\cite{huaiSNAILRadarLargescale2025} at 1~Hz.
Revisits require more than 20~s separation, at most 10~m spatial distance, and at least 50~m of reference-path travel, yielding 14,764 candidate pairs.
Lidar scans are motion-compensated to their scan timestamps using per-point acquisition times and local reference motion.
For plotting, both estimates must succeed, camera viewing directions must differ by at most $90^\circ$, and
$\|\mathbf t^{\mathrm{lidar}}_{i\leftarrow j}-\mathbf t^{\mathrm{stereo}}_{i\leftarrow j}\|_2\leq1$~m.
Stereo estimates additionally satisfy $e_t\leq5$~m and $e_R\leq10^\circ$.

\begin{figure}[!p]
\centering
\subfloat[]{\includegraphics[width=0.72\linewidth]{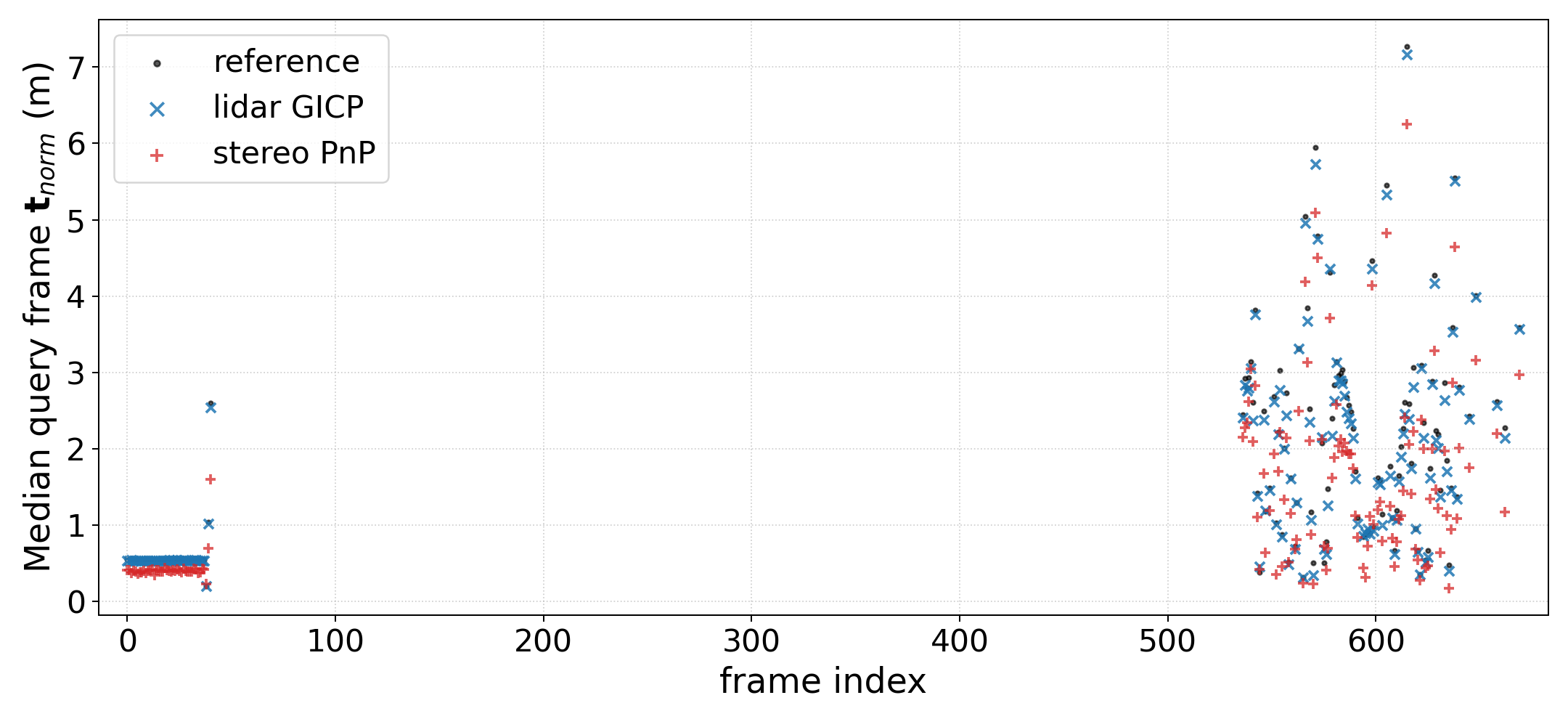}}\\[-0.4ex]
\subfloat[]{\includegraphics[width=0.72\linewidth]{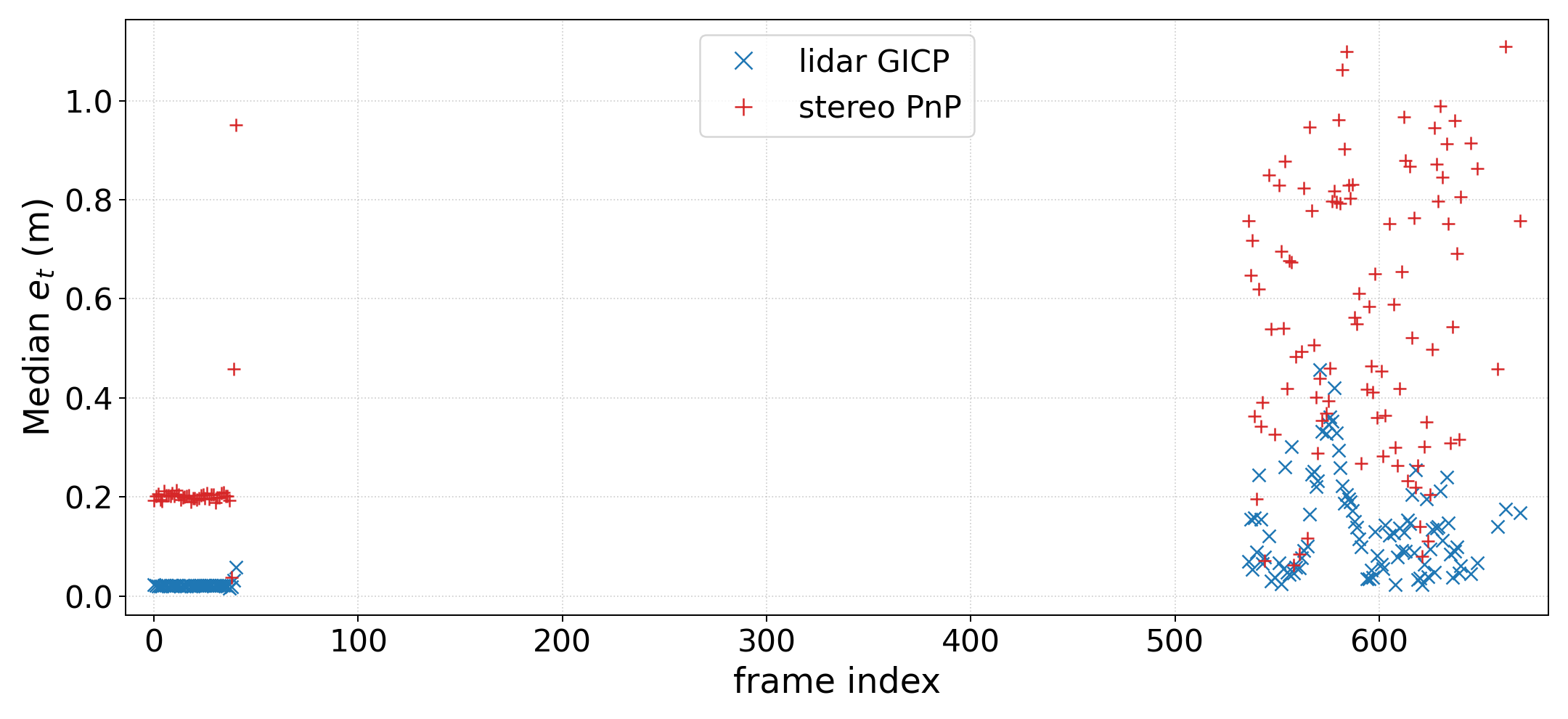}}\\[-0.4ex]
\subfloat[]{\includegraphics[width=0.72\linewidth]{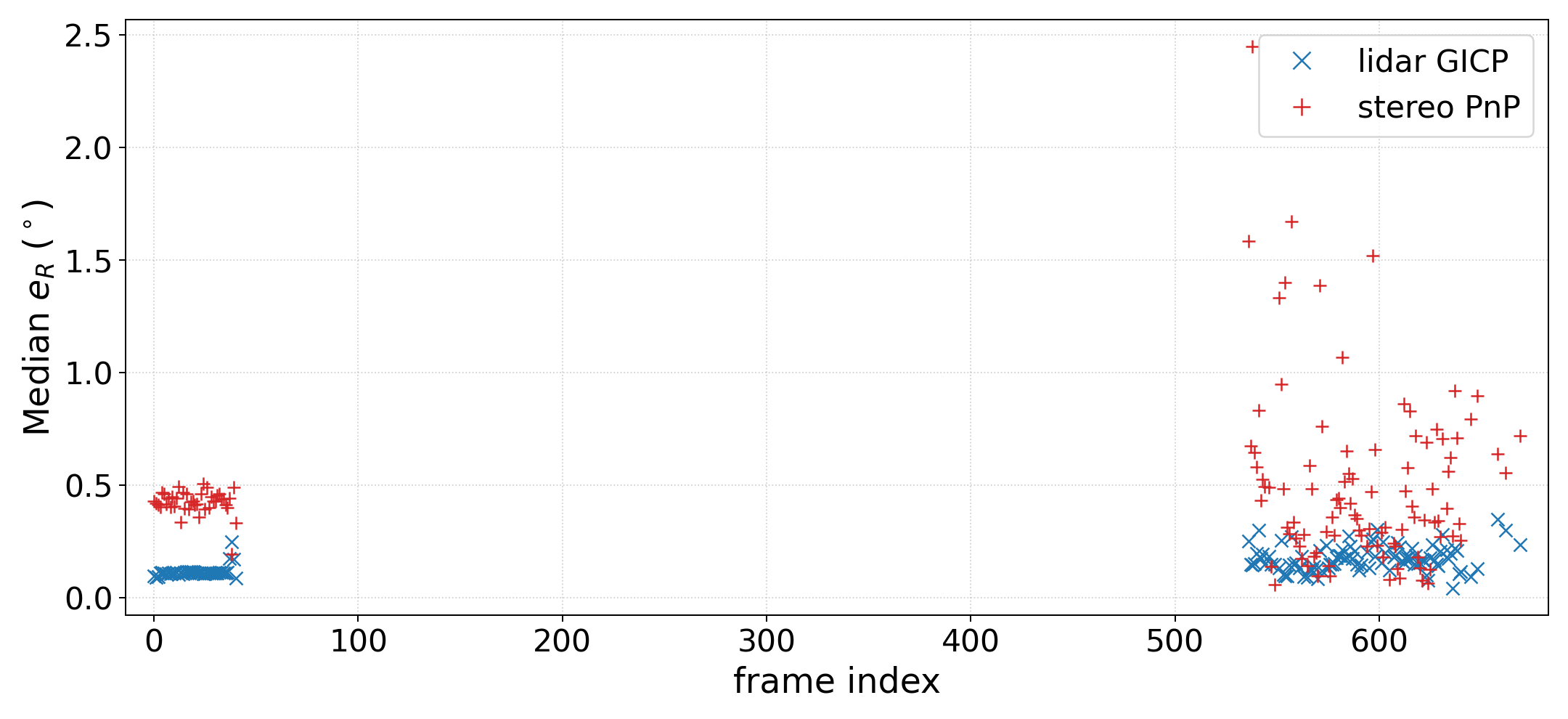}}\\[-0.4ex]
\subfloat[]{\includegraphics[width=0.44\linewidth]{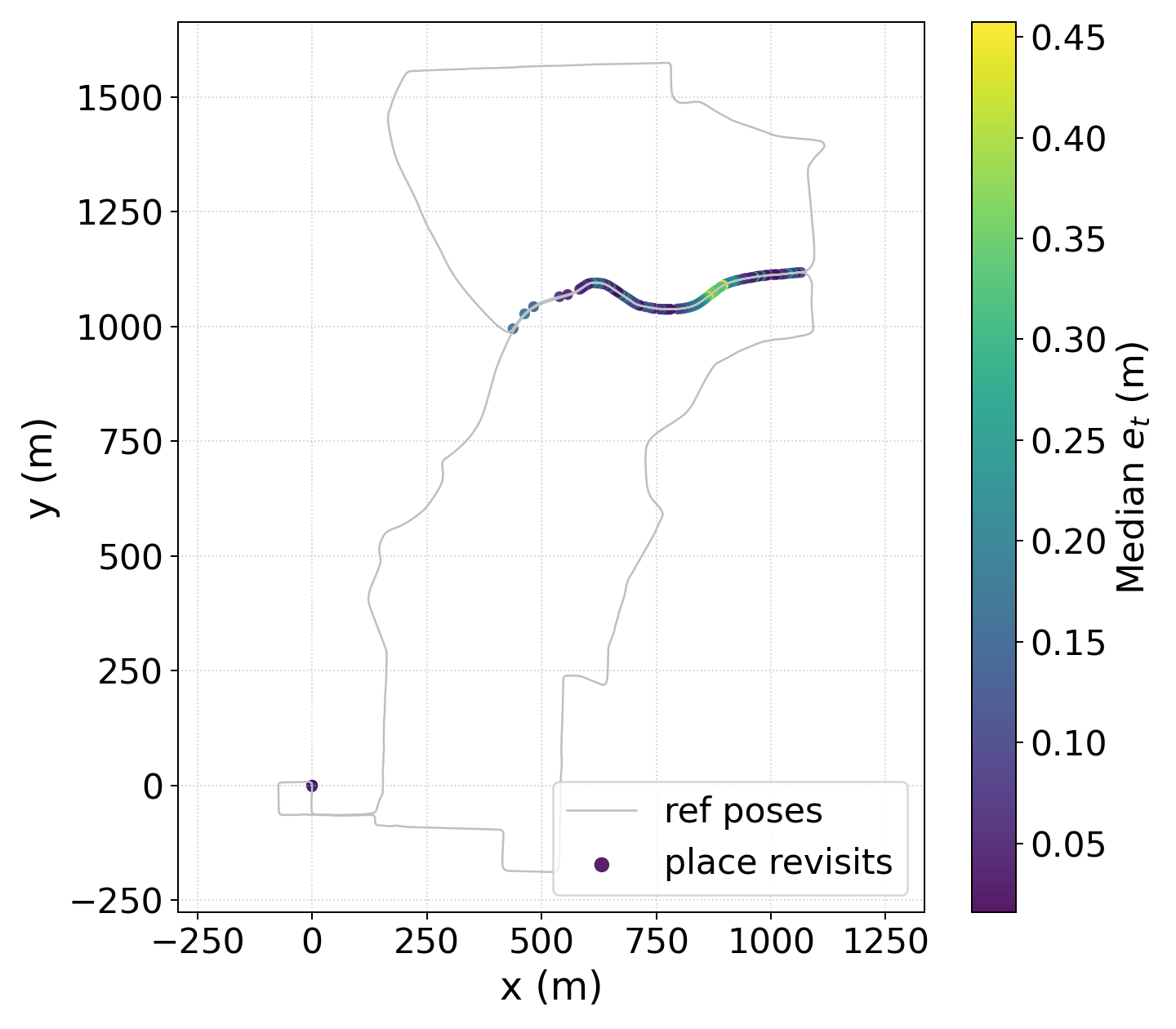}}
\caption{SNAIL Radar 20240113/5 diagnostics on revisit pairs.
(a) Relative translation magnitudes using the same pair for reference, lidar, and stereo, selected by median lidar translation discrepancy for each train frame.
(b,c) Translation and rotation discrepancies for median representatives.
(d) Lidar translation discrepancy from (b) at train-frame locations along the reference trajectory.}
\label{fig:snail20240113_5}
\end{figure}

Fig.~\ref{fig:snail20240113_5} shows sequence 20240113/5, for which 1,197 of 2,002 pairs remain, covering 139 train frames.
The maximum plotted lidar query median discrepancies are 0.46~m and $0.35^\circ$; stereo estimates exhibit slightly larger scatter, and the trajectory view localizes the remaining deviations.
These results support local revisit consistency of the SNAIL radar full reference trajectories.

\section{Conclusion}
This paper presented a place-revisit consistency evaluation method for reference poses in mobile sensing benchmarks. Applied to KITTI Odometry, the method identified likely local reference-pose inconsistencies, including median translation discrepancies up to 6.60 m. The agreement between independent sensor-derived estimates, the temporal consistency of lidar estimates across nearby train frames, and manual point-cloud overlays support the conclusion that the highlighted revisit regions deserve careful treatment in benchmark-based evaluation.
The SNAIL Radar test shows revisit consistency using motion-compensated lidar scans and stereo images.
Future work will package the analysis as a reproducible validation script and extend it to further mobile sensing datasets.

\clearpage
\bibliographystyle{IEEEtran}
\bibliography{zotero}

\end{document}